\documentclass[11pt]{article}

\usepackage[final]{acl}

\usepackage{times}
\usepackage{latexsym}
\usepackage{graphicx}
\usepackage{amsmath}
\usepackage{amssymb}
\usepackage{booktabs}
\usepackage{mathalfa}
 \usepackage{microtype}
\usepackage{times}
\usepackage{latexsym}

\usepackage[T1]{fontenc}

\usepackage[utf8]{inputenc}

\usepackage{microtype}

\usepackage{inconsolata}
 \usepackage{natbib} 
\usepackage{amssymb}%
\usepackage{pifont}%
\usepackage{graphicx}
\usepackage{hyperref}
\usepackage{url}
\usepackage{microtype}
\usepackage{times}
\usepackage{latexsym}
\usepackage{soul}
\usepackage{amsmath}
\usepackage{array, booktabs}
\usepackage{bm}
\usepackage[capitalize]{cleveref}
\usepackage{graphicx}
\usepackage{tabularx}
\usepackage{soul}
\usepackage{xcolor}
\usepackage{todonotes}
\usepackage{multirow}
\usepackage{xpatch}
\usepackage{blindtext}
\usepackage{fdsymbol}
\usepackage{microtype}
\usepackage{hyperref}
\usepackage[export]{adjustbox}
\usepackage{textcomp}
\usepackage{xparse}
\usepackage{enumitem}
\usepackage{makecell}
\usepackage{subcaption}
\usepackage{colortbl}
\usepackage{tcolorbox}
\usepackage{xparse}
\usepackage{stfloats}
\usepackage{arydshln}
\usepackage{bbm}
\usepackage[linesnumbered,vlined,ruled]{algorithm2e}
\usepackage{url}
\usepackage{graphicx}
\usepackage{xspace}
\usepackage{colortbl}
\usepackage{booktabs}
\usepackage{multirow}
\usepackage{xspace}

\usepackage{fancyhdr}
\fancypagestyle{firstpage}{%
  \fancyhf{}
  \fancyhead[C]{\textcolor{red}{\textbf{Published in the Findings of the Association for Computational Linguistics: ACL 2026}}}
  
}

\definecolor{ForestGreen}{rgb}{0, 0.69, 0.31}
\definecolor{NavyBlue}{rgb}{0, 0.44, 0.75}
\newcommand{\hgreen}[1]{\textcolor{ForestGreen}{\textbf{#1}}} 

\usepackage{pifont}

 \newcommand{\methodname}{S2H-DPO\xspace}
\definecolor{Cerulean}{rgb}{0.0, 0.48, 0.65}

\definecolor{c2}{RGB}{218,0,0}

\definecolor{lightblue}{RGB}{212, 235, 255}
\definecolor{lightorange}{RGB}{255, 204, 168}
\definecolor{lightyellow}{RGB}{255, 255, 168}
\definecolor{lightred}{RGB}{255, 168, 168}
\definecolor{darkred}{RGB}{234, 107, 102}
\definecolor{darkerblue}{RGB}{103, 136, 184}
\definecolor{lightgreen}{RGB}{102, 205, 102}

\definecolor{gold}{rgb}{0.83, 0.69, 0.22}
\sethlcolor{lightblue}
\newcolumntype{Y}{>{\centering\arraybackslash}X}

\NewDocumentCommand{\steeve}
{ mO{} }{\textcolor{gold}{\textsuperscript{\textit{Steeve}}\textsf{\textbf{\small[#1]}}}}
\NewDocumentCommand{\jason}
{ mO{} }{\textcolor{red}{\textsuperscript{\textit{Jason}}\textsf{\textbf{\small[#1]}}}}

\NewDocumentCommand{\shafiq}
{ mO{} }{\textcolor{blue}{\textsuperscript{\textit{Shafiq}}\textsf{\textbf{\small[#1]}}}}
\NewDocumentCommand{\sid}
{ mO{} }{\textcolor{green}{\textsuperscript{\textit{Sid}}\textsf{\textbf{\small[#1]}}}}
\NewDocumentCommand{\philippe}
{ mO{} }{\textcolor{magenta}{\textsuperscript{\textit{Philippe}}\textsf{\textbf{\small[#1]}}}}
\NewDocumentCommand{\can}
{ mO{} }{\textcolor{violet}{\textsuperscript{\textit{Can}}\textsf{\textbf{\small[#1]}}}}
\NewDocumentCommand{\caiming}
{ mO{} }{\textcolor{orange}{\textsuperscript{\textit{Caiming}}\textsf{\textbf{\small[#1]}}}}
\NewDocumentCommand{\ap}
{ mO{} }{\textcolor{brown}{\textsuperscript{\textit{Akshara}}\textsf{\textbf{\small[#1]}}}}

\definecolor{gold}{rgb}{0.83, 0.69, 0.22}

\newcommand{\mathv}[1]{\textsc{Math-Vision}}
\newcommand{\chocolate}[1]{\textsc{Chocolate}}

\crefformat{section}{\S#2#1#3} %
\crefformat{subsection}{\S#2#1#3}
\crefformat{subsubsection}{\S#2#1#3}

\usepackage{xcolor}
\usepackage[T1]{fontenc}

\usepackage[utf8]{inputenc}

\usepackage{microtype}

\usepackage{inconsolata}

\usepackage{graphicx}

\title{\methodname: Hardness-Aware Preference Optimization for Vision–Language Models}

\author{
 \textbf{Nitish Shukla\textsuperscript{1}}
 \textbf{Surgan Jandial\textsuperscript{2}}
 \textbf{Arun Ross\textsuperscript{1}}
\\
\\
 \textsuperscript{1}Michigan State University
 \textsuperscript{2}Carnegie Mellon University
\\
 \small{
   \textbf{Correspondence:} \texttt{\href{mailto:email@domain}{shuklan3@msu.edu}}
 }
}

\begin{document}
\maketitle
\thispagestyle{firstpage}
\begin{abstract}

Vision-Language Models (VLMs) have demonstrated remarkable progress in single-image understanding, yet effective reasoning across multiple images remains challenging. We identify a critical capability gap in existing multi-image alignment approaches: current methods focus primarily on localized reasoning with pre-specified image indices (``Look at Image 3 and...''), bypassing the essential skills of global visual search and autonomous cross-image comparison. To address this limitation, we introduce a Simple-to-Hard (S2H) learning framework that systematically constructs multi-image preference data across three hierarchical reasoning levels  requiring an increasing level of capabilities: (1) single-image localized reasoning, (2) multi-image localized comparison, and (3) global visual search. Unlike prior work that relies on model-specific attributes, such as hallucinations or attention heuristics, to generate preference pairs, our approach leverages prompt-driven complexity to create chosen/rejected pairs that are applicable across different models. Through extensive evaluations on LLaVA and Qwen-VL models, we show that our diverse multi-image reasoning data significantly enhances multi-image reasoning performance, yielding significant improvements over baseline methods across benchmarks. Importantly, our approach maintains strong single-image reasoning performance while simultaneously strengthening multi-image understanding capabilities, thus advancing the state of the art for holistic visual preference alignment.

\end{abstract}

\section{Introduction}

\label{sec:intro}
Vision–Language Models (VLMs) \cite{LLaVAnextinterleave,qwenvl,videoLLaVA} have made rapid progress, yet their performance drops for challenging tasks and inputs. In particular, we study reasoning of VLMs across multiple images where the model must not only interpret each image in isolation, but also align, compare, and integrate evidence across images. While proprietary systems such as GPT-4o \cite{2024gpt4o} demonstrate strong multi-image capabilities, open-source VLMs \citep{LLaVAnextinterleave,xcomposer2d5} still struggle to reliably aggregate information when visual context is distributed across several images. The difficulty compounds as the number of images grows, increasing both the search space and the need for consistent cross-image correspondence. Effective multi-image reasoning therefore hinges on two core abilities: (1) localizing where to look across multiple images, and (2) composing information from multiple regions into a coherent conclusion.
\begin{figure}
    \centering
    \includegraphics[width=\linewidth]{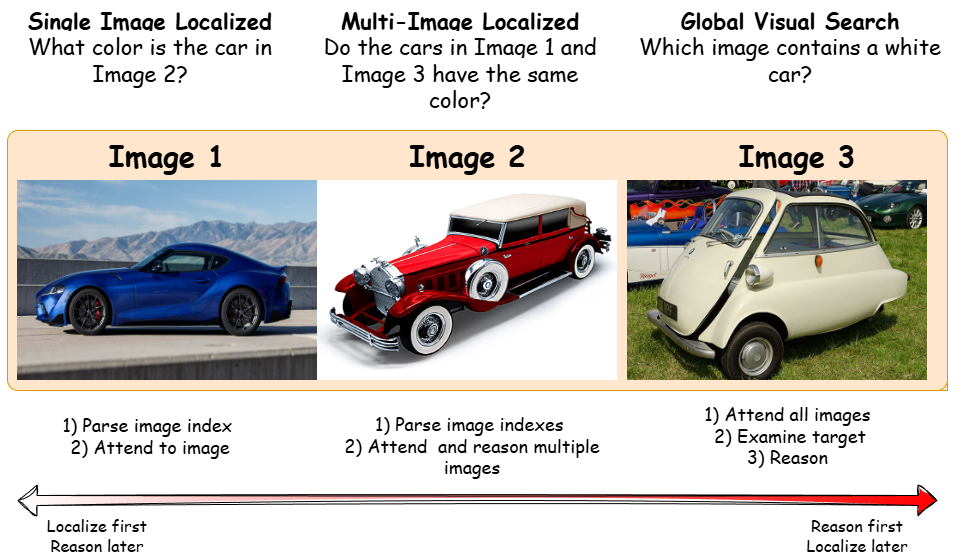}
    \caption{Multi-image reasoning skill hierarchy: From localized reasoning ($L$1: "What is   the color of the car in Image 2?") to global visual search ($L$3: "Which image contains a white car?"). Each level requires strictly more capabilities than the previous.}
    \label{fig:cars-comp}
\end{figure}
\begin{figure*}
    \centering
    \includegraphics[width=\linewidth]{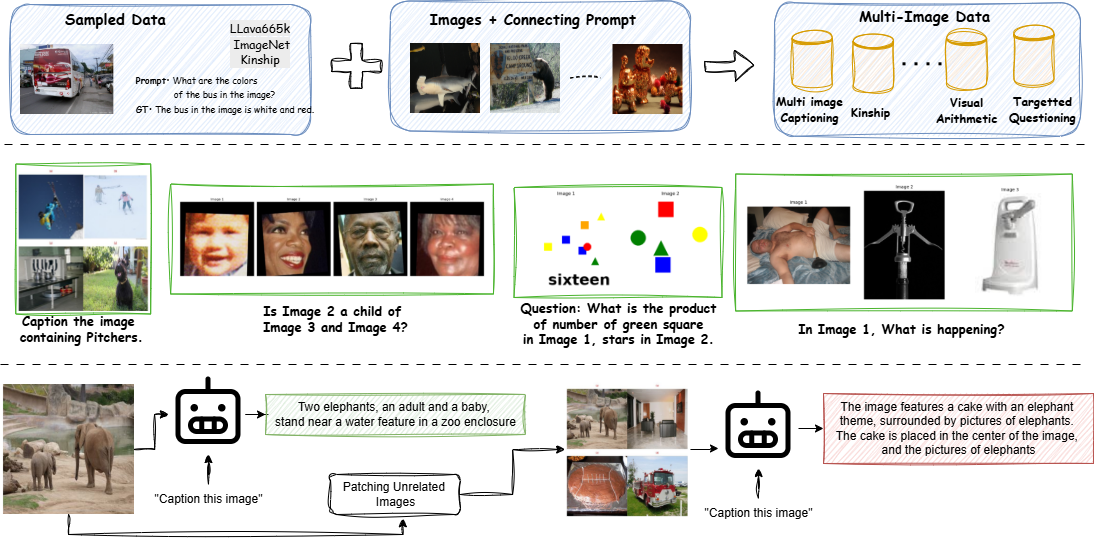}
    \caption{\textbf{Overview of our method.} We transform single-image data sources (e.g., LLaVA 665k, ImageNet) into multi-image datasets across visual modalities with progressively increasing cognitive load. Our synthetic transformation pipeline creates task hierarchies that advance from basic single-image understanding to complex multi-image reasoning requiring visual comparison, spatial reasoning, and cross-modal integration. }
    
    \label{fig:overview}
\end{figure*}
Existing work primarily addresses this via multi-image pre-training \cite{flaminggo} and supervised fine-tuning (SFT) \cite{mantis,chen2024sharegpt4video,liu2024mmdu}, with only a few studies \cite{liu2025miadpo} approaching the problem from the perspective of visual preference alignment. In particular, \cite{liu2025miadpo} observes that simply incorporating multi-image data during SFT is not sufficient to improve multi-image reasoning—and can, in some cases, even degrade performance on single-image tasks—thereby motivating Direct Preference Optimization (DPO)-style preference alignment for multi-image reasoning. However, a key challenge is constructing multi-image preference pairs because (1) each example must include a carefully selected set of multiple images and corresponding questions, and (2) each example requires well-curated chosen and rejected text response. These requirements make large-scale human annotation prohibitively expensive, motivating synthetic data approaches for generating multi-image preference data.

While recent work—most notably MIA-DPO \cite{liu2025miadpo}—proposes promising strategies for generating synthetic preference pairs, its data curation remains fundamentally limited in capturing the skills required for robust multi-image reasoning. In particular, it relies on questions that pre-specify where the model should look (e.g., “What is the color of the car in Image 2?” given multiple images). By explicitly indicating which image(s) to examine, this formulation sidesteps key competencies in which the model must autonomously determine relevant visual evidence, compare images, and compose information across multiple views. Therefore, we argue that, beyond simply reducing the cost of large-scale data generation, it is crucial to explicitly define the capabilities needed for multi-image reasoning and to ensure that data curation targets them. In this regard, let us consider the progression of multi-image reasoning capabilities:
\begin{itemize}[leftmargin=*, itemsep=0pt]
\item \textbf{Level 1 (Single-Image Localized):} "What is the color of the car in Image 2?" — Reason about one pre-specified image. 
\item \textbf{Level 2 (Multi-Image Localized):} "Do the cars in Image 1 and Image 3 have same color?" —  Reason and Compare across multiple pre-specified images, unlike Level 1 which reduces to single-image reasoning.
\item \textbf{Level 3 (Global Visual Search):} "Which images contain a white car?" — Reason and Search all images to locate relevant regions and carefully aggregate information, unlike Level 1 and Level 2, where regions are pre-specified.
\end{itemize}

Figure~\ref{fig:cars-comp} illustrates these capabilities, where each level demands strictly more from the model than the previous level. Critically, MIA-DPO \cite{liu2025miadpo} trains only on Level~1. We argue this is not merely a difference in question format, but an important gap: different formulations induce qualitatively different reasoning patterns with increasing difficulty. In particular, at Level~1 the model reasons about a single, pre-specified image (``look at Image~2''); at Level~2 it must compare multiple pre-specified images (``look at Image~1 and Image~3''); and at Level~3 it must autonomously determine where to look and then reason over the relevant regions (e.g., ``look for the image with a white car''). Thus, we ask the question: \textbf{Can preference tuning on diverse multi-image reasoning tasks achieve broader generalization than index-based localization training alone?}

More precisely, we automatically construct a diverse synthetic training set containing preference pairs spanning Levels 1-3, allowing models to learn multi-image reasoning capabilities simultaneously. We then perform a \textbf{comprehensive multi-image DPO approach that jointly trains models across all reasoning levels}. Through systematic ablations, we demonstrate that: (1) joint training on mixed-difficulty data enables robust multi-image reasoning, and (2) models trained on diverse question types, including global search, generalize better than those trained only on localized questions.

Experimental results highlight a significant capability gap between our approach and the current state-of-the-art. While MIA-DPO demonstrates moderate accuracy on lower-complexity benchmarks, its performance notably declines when faced with more challenging Level 3 tasks. Conversely, our method—which leverages training samples across all reasoning levels—consistently outperforms MIA-DPO on both tiers, with a more pronounced advantage on the harder tasks (see Table \ref{multi-image-benchmarks}). A critical limitation of MIA-DPO is its reliance on Level 1 examples, which hinge on model-specific intrinsic hallucinations to generate chosen-rejected pairs. This necessitates the generation of a new dataset for every new model architecture. By introducing higher hardness levels through prompt-driven complexity, our method alleviates this dependency. Furthermore, our approach preserves single-image reasoning capabilities, matching the performance of the pre-finetuned baseline. These findings demonstrate that our method substantially enhances multi-image reasoning across all semantic levels without compromising foundational single-image performance.

In summary, our main contributions are:
\begin{itemize}[leftmargin=*, itemsep=0pt]
\item We identify a ``capability gap'' in existing multi-image alignment: current methods train only on localized reasoning with pre-specified images (Level 1), missing the global and partial visual search capabilities (Levels 2-3) required for real-world multi-image understanding.

\item We develop a systematic technique for generating high-quality rejected answers from correct-answer-only datasets, enabling comprehensive multi-image preference optimization across all reasoning levels without manual annotation.

\item Through extensive experiments, we demonstrate that our comprehensive multi-image DPO approach substantially outperforms existing methods while preserving single-image reasoning capabilities.
\end{itemize}

\section{Proposed Method}

\subsection{Preliminaries }
In this section, we present the concept of visual preference alignment and use the Direct Preference Optimization (DPO) method as a representative example.

\paragraph{Visual Preference Alignment:}
Preference alignment focuses on aligning a model’s outputs with a set of preferences. Common methods include \textbf{R}einforcement \textbf{L}earning from \textbf{H}uman \textbf{F}eedback (\textbf{RLHF}) \cite{yu2024rlhfv} and \textbf{R}einforcement \textbf{L}earning from \textbf{AI} \textbf{F}eedback (\textbf{RLAIF})\cite{yu2024rlaif}. Consider a dataset \(D\) where each sample consists of an input prompt \(x\), a preferred response \(y_w\), and a rejected response \(y_l\). Formally, we represent the dataset as $D = \{x, y_w, y_l\}.$
The input \(x\) can be an interleaved sequence of images \(v\) and text \(t\).

When a VLM processes \(x\) to generate an output \(y\), a reward \(r(x, y)\) is assigned by a reward model \(r\), which assigns higher scores to preferred outputs and lower scores to rejected ones. Visual preference alignment aims to maximize this reward:

$
\max_{\theta} \mathbb{E}_{x \sim D, y \sim \pi_{\theta}(y|x)} \left[r(x, y)\right],
$
where $\theta$, $\pi_{\theta}$ and $\pi_{\theta}(y|x)$ denote the parameters, policy, and output distribution of VLM, respectively.

To avoid overfitting on the dataset $D$,  a KL-divergence loss $D_{\text{KL}}$ is incorporated by preference alignment approaches to regularize the difference between the model's policy $\pi_{\theta}(y|x)$ and a reference model's policy $\pi_{\text{ref}}(y|x)$:
\begin{equation}
\resizebox{\columnwidth}{!}{$
    \max_{\theta} \left[ \mathbb{E}_{x \sim D, y \sim \pi_{\theta}(y|x)} \left[r(x, y)\right] - \beta \cdot D_{\text{KL}}(\pi_{\theta}(y|x) \parallel \pi_{\text{ref}}(y|x)) \right],
    $}
\label{eq:reward}
\end{equation}
where, the hyper-parameter $\beta$ controls the influence of KL-divergence on the optimization objective. Note that the reference model is the model's state prior to preference alignment.

\paragraph{Direct Preference Optimization (DPO):}
To optimize the preference alignment objective in Eq.~(\ref{eq:reward}), we can use either an online reward model (\textit{e.g.}, PPO \cite{schulman2017proximal}) or pre-computed off-line chosen/rejected pairs (\textit{e.g.}, DPO \cite{rafailov2024direct}).
Given its simplicity, DPO has been widely adopted in previous visual alignment works \cite{zhao2024beyond,zhou2024aligning}.
Eq.~(\ref{eq:reward}) can be  reformulated  as the loss function of DPO: $L_{\mathrm{DPO}}(\pi_\theta; \pi_{\mathrm{ref}}) =$
\begin{equation}
\resizebox{\columnwidth}{!}{$
- \mathbb{E}_{(x, y_w, y_l) \sim D} 
\Big[ \log \sigma \big(
\beta \log \frac{\pi_\theta(y_w \mid x)}{\pi_{\mathrm{ref}}(y_w \mid x)} 
- \beta \log \frac{\pi_\theta(y_l \mid x)}{\pi_{\mathrm{ref}}(y_l \mid x)}
\big) \Big],
$}
\label{eq:dpo_losse}
\end{equation}
\noindent where, $\sigma(.)$ denotes the sigmoid function. As shown in Eq.~(\ref{eq:dpo_losse}), DPO-based alignment methods concentrate on designing input prompts \(x\) and pairing each with a preferred output \(y_w\) and a rejected output \(y_l\).
\begin{figure*}
    \centering
    \includegraphics[width=0.9\linewidth]{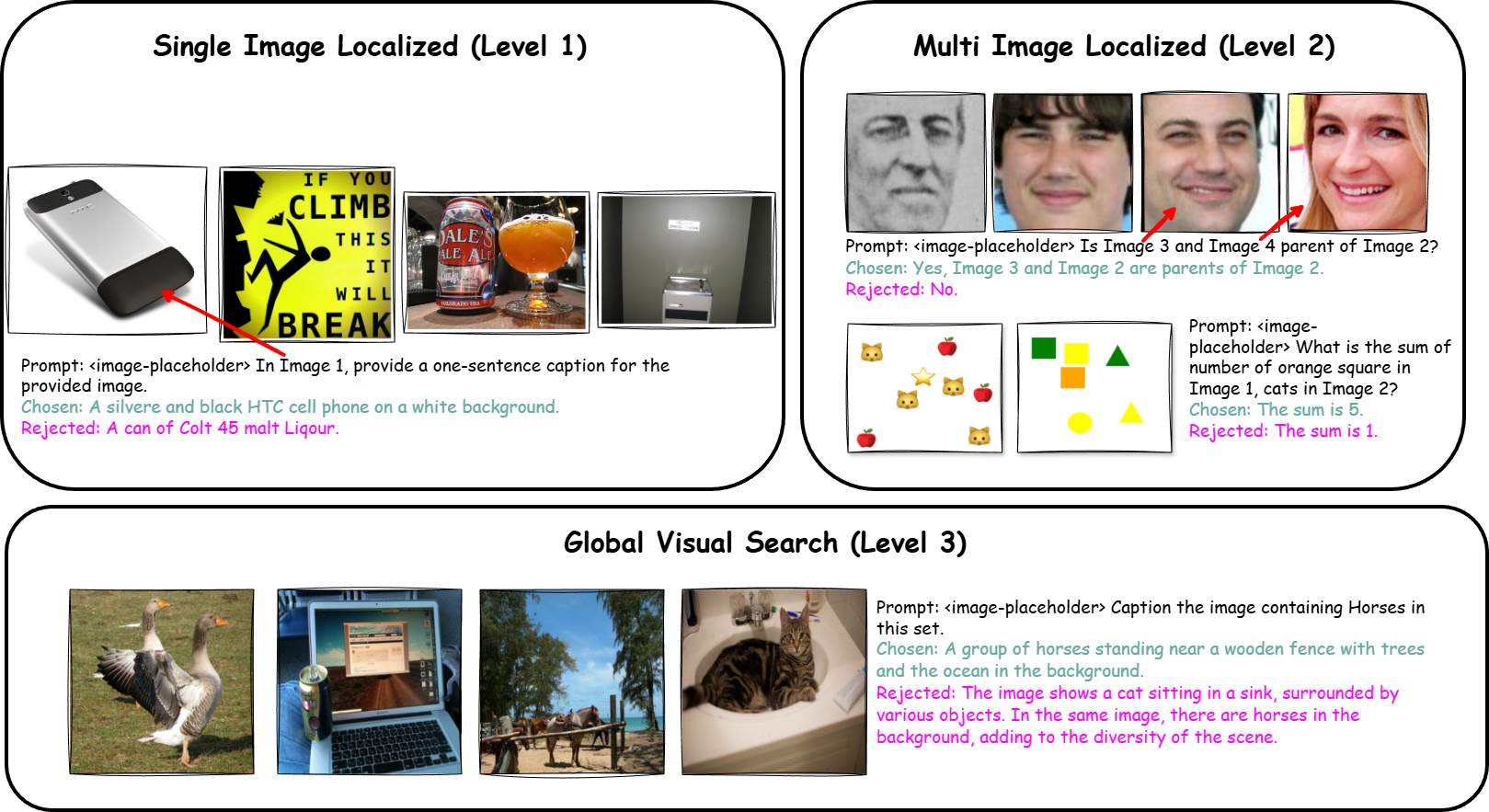}
\caption{Simple-to-Hard (S2H) DPO Data Format. \textbf{Level 1:} The query directly points to a specific image (\emph{e.g.}, ``Caption the first image''), and rejected pairs provide responses completely unrelated to the target image. \textbf{Level 2:} The query references multiple images to enforce multi-image reasoning (\emph{e.g.}, ``Compare the first and third images''). \textbf{Level 3:} The query is open-ended and requires the model to examine all images before identifying which one satisfies the semantic constraint (\emph{e.g.}, ``Caption the image containing a peacock'').}
\label{fig:data}
\end{figure*}

\subsection{Simple-to-hard (S2H) Tasks}
We propose three Simple-to-Hard (S2H) probe tasks with increasing levels of complexity, defined by the cognitive load required for localization, cross-image grounding, and reasoning. Our tasks therefore escalate from basic single-image query to complex multi-step reasoning that demands joint understanding of all images. We illustrate the overall process in Figure \ref{fig:overview}. Below, we explain these tasks in detail. 

\subsubsection{Single Image Localized (Level 1)}
In the simplest case, following prior work~\cite{liu2025miadpo}, we convert single-image datasets to multi-image format by appending unrelated distractor images to existing VQA samples. The chosen response corresponds to the gold answer for the target image, while the rejected response is an incorrect answer generated by the model. Critically, this approach does not establish explicit relationships between images—the task simply requires the model to ignore irrelevant visual context. The rejected responses primarily arise from model hallucinations when processing noisy multi-image inputs, rather than from meaningful cross-image reasoning errors. While this provides a baseline for multi-image handling, it does not test genuine multi-image reasoning capabilities.
\subsubsection{Multi-Image Localized (Level 2)}
In this category, we extend Level 1 tasks to a multi-image setting. Rather than analyzing a single image in isolation, we expand the context by requiring the model to localize features across multiple images and respond to prompts that necessitate explicit cross-image connections. We introduce two distinct tasks in this category: (i) Kinship Recognition, where the objective is to determine familial or social relationships within a set of facial images; and (ii) Visual Arithmetic, which requires the model to perform mathematical reasoning—such as counting or logical operations—on shapes, numbers, and objects distributed across several images. We select Kinship Recognition and Visual Arithmetic as Level-2 tasks because they capture complementary aspects of multi-image localized reasoning. Kinship Recognition requires asymmetric relational inference: although multiple images are presented, the answer pertains primarily on a single target image in relation to others, demanding cross-image relational grounding. Visual Arithmetic, in contrast, requires symmetric compositional aggregation, where each image contributes equally and the model must combine information across all images to compute the answer. Together, these sub-skills capture the core challenges of L2 multi-image reasoning.  

\textit{Chosen/Rejected pair generation.} For Kinship Recognition, we leverage preexisting datasets~\cite{kinFG2017,robinson2016families} to generate chosen and rejected pairs. The process is straightforward: we first read the labels and relationships of the individuals. If the relationship is correct, we generate a chosen deterministic caption (e.g., ‘Yes, Image 3 is…’) and a corresponding rejected caption (e.g., ‘No, Image 3 is not…’). Conversely, if the relationship is incorrect, we generate a chosen–rejected pair by flipping the responses accordingly.  For Visual Arithmetic, images are generated synthetically using deterministic procedures, ensuring that each image contributes clearly to the final answer while maintaining controlled, reproducible inputs for multi-image reasoning. Images are generated synthetically by placing random shapes and objects in each image while keeping track of their counts. To create a question, we randomly sample a subset of images, select some objects, and choose a random mathematical operator (e.g., addition, subtraction) to form a problem whose answer depends on aggregating information across the selected images. This deterministic procedure guarantees correct chosen/reject pair generation.

\begin{figure*}
    \centering
    \includegraphics[width=0.9\linewidth]{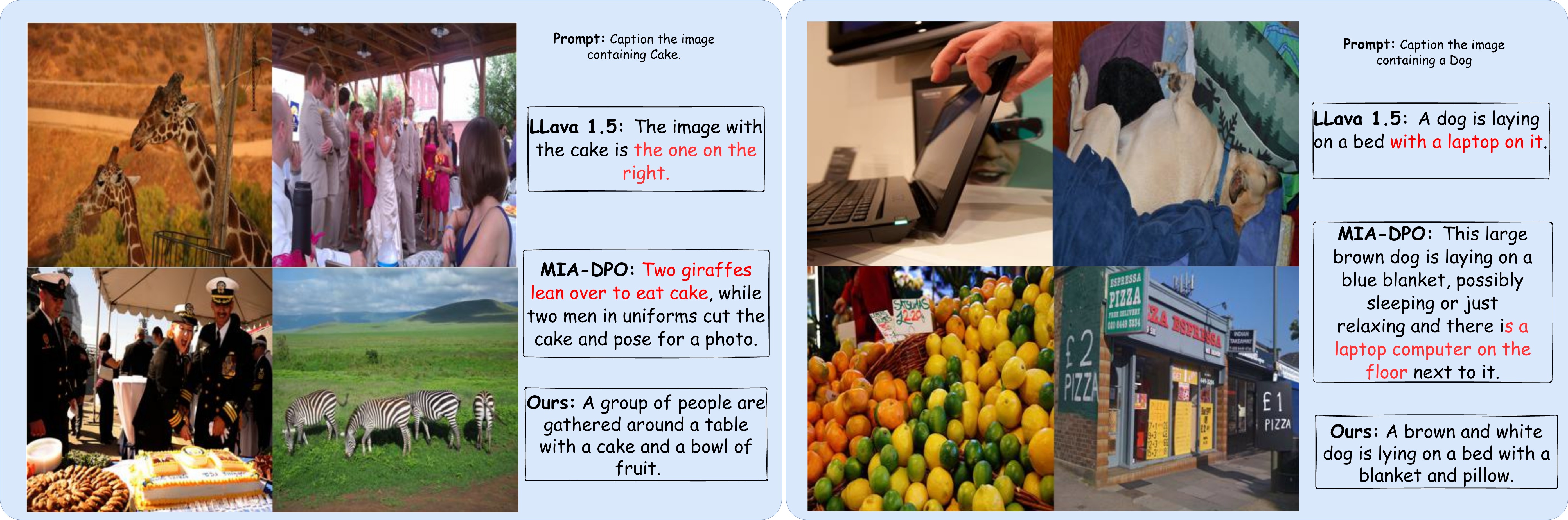}
    \caption{Outputs of our method on Global Visual Search. Our approach effectively isolates the queried concept and produces accurate responses to the prompt. }
    \label{fig:outputs}
\end{figure*}

\subsubsection{Global Visual Search (Level 3)} 
This category inverts the computational ordering: the model must first perform global reasoning across the entire image set before localizing the relevant image. Consider the task of captioning an image that satisfies a semantic criterion, \emph{e.g.}, ``Caption the image containing a peacock''. Unlike previous levels, localization cannot precede reasoning. Instead, the model must evaluate all images against the query constraint, identify the matching instance, and only then generate the caption. This formulation tests the model's ability to integrate cross-image search with subsequent generative tasks.

\noindent\textbf{Single-to-Multi-Image Prompt Construction.} 
Figure~\ref{fig:data} (bottom) illustrates our procedure for converting single-image captions into multi-image chosen-rejected pairs. We construct training data from existing annotated datasets through a two-stage process designed to distinguish targeted search from general summarization.
\textit{Chosen pair generation:} We first select a target concept from ImageNet \cite{5206848} and sample a target image $\mathcal{I}_{\text{target}}$ belonging to that concept. We then prompt a vision-language model to generate both (1) a detailed caption and (2) the primary object present in the image (\emph{e.g.}, a peacock). To form a multi-image set, we pair $\mathcal{I}_{\text{target}}$ with $N-1$ distractor images randomly sampled from \emph{different} concept classes, ensuring that the same primary object does not appear in more than one image. The chosen response is then assigned as the generated caption.
\textit{Rejected pair generation:} For the same multi-image set, we prompt the captioning model \textit{without} specifying a target object, asking it to simply ``caption the images.'' This yields a rejected response that typically either aggregates information across multiple images or provides a generic description, failing to isolate the target. Importantly, these rejected responses are not randomly unrelated—they represent plausible but incorrect responses, which provides a better learning signal to model compared to trivial negatives. We use Qwen2.5-VL-32B \cite{bai2025qwen2} as the caption generator in this work.
\textit{Quality filtering:} To ensure meaningful contrast between chosen and rejected pairs, we compute semantic similarity between the two responses using text encoders (CLIP, MPNet). Pairs with similarity exceeding threshold $\tau$ (top quartile) are discarded, as high similarity indicates the rejected response inadvertently captured the correct behavior. This filtering ensures the contrastive pairs provide a clear learning signal for distinguishing conditional visual search from unconstrained multi-image description.

\subsection{Implementation Details}
Our \methodname is model-agnostic and can be directly applied to a wide range of LVLMs without requiring the generation of additional training datasets, in contrast to prior approaches. We evaluate our method on three widely used MLLMs: LLaVA-v1.5-7B~\cite{liu2024improved} and Qwen2.5-VL-7B~\cite{bai2025qwen2} and Qwen3-VL-2B~\cite{bai2025qwen3}. All models are trained for three epochs with a learning rate of $5\times10^{-5}$, and the temperature parameter $\beta$ is set to 0.1. We create 20K samples for each level.

\section{Related Works in Preference Alignment}
Aligning Large Language Models with human preferences has become a critical requirement for their safe and reliable deployment in real-world applications \cite{misaligned_llm, llm_faithful_cot, ai_safety, llm_sens, prompt_safety}, particularly under stringent ethical and safety constraints. Preference alignment methods can be broadly divided into two categories. The first involves feedback-driven alignment, which leverages either human-annotated preferences~\citep{rafailov2024direct,bai2022training} or AI-generated feedback~\citep{yu2024rlaif}. The second category focuses on prompt-based guidance, where carefully designed instructions are used to steer model behavior without modifying model parameters~\citep{chen-etal-2023-chatcot}.

Similarly, recent works investigate the unreliabilty and misaligned behavior of vision--language large models (VLLMs) across various tasks \cite{fine_web, gui_ground, more_images, vlm_blind}, motivating the development of specialized vision-language preference alignment approaches. ~\citep{sun2023aligning} introduced LLaVA-RLHF, which utilizes human-annotated preference data to reduce hallucinations in LLaVA. \cite{li-etal-2024-vlfeedback} proposed a preference distillation framework that transfers alignment signals into VLLMs, improving visual grounding and response relevance. Similarly, ~\cite{yu2024rlhfv} collected fine-grained human preferences in the form of segment-level corrections to hallucinated content and optimized model behavior using dense supervision. HA-DPO \cite{zhao2024beyond} addresses alignment by leveraging GPT-4’s API to create the required DPO data, though it incurs substantial API expenses. POVID \cite{zhou2024aligning} induces hallucinations using blurred images GPT-4 to generate DPO data. Recently, MIA-DPO \cite{liu2025miadpo} automates the generation of chosen-rejected pairs by exploiting the model’s intrinsic hallucinations, but it still necessitates creating a fresh set of DPO data for each model.
\section{Experiments}

\subsection{Benchmarks}
To assess our method, we evaluate its performance across two primary categories of benchmarks. First, we examine multi-image reasoning using three multi-image benchmarks BLINK \citep{blink}, MANTIS \citep{mantis}  and NLVR2 \citep{nlvr2}. 
Second, to ensure our approach maintains robust single-image capabilities, we test on two single-image benchmarks, including MMStar \citep{mmstar} and  POPE \citep{pope}. This extensive evaluation demonstrates our method's versatility and confirms significant performance gains in multi-image scenarios without sacrificing single-image proficiency.

\subsection{Baseline Methods} We benchmark our method against several state-of-the-art preference optimization strategies. LLaVA-RLHF \citep{sun2023aligning} serves as the standard RLHF baseline, utilizing human and GPT-generated feedback. HA-DPO \citep{zhao2024beyond} focuses on error correction by leveraging GPT-4V as an expert annotator to rectify model hallucinations. In contrast, POVID \citep{zhou2024aligning} adopts an adversarial approach, pairing distorted images with synthetic hallucinations to stress-test modality alignment. Finally, we compare against MIA-DPO \citep{liu2025miadpo}, which automates the generation of preference data by exploiting the model’s  hallucinations in multi-image contexts.

\begin{table}[ht]
\caption{
\small
\textbf{Main results on multi-image benchmarks.} We compare our \methodname with other DPO algorithms across five multi-image benchmarks. Our method yields consistent improvements over both the classic LLaVA-v1.5 and the recent Qwen2.5-VL-7B. In contrast, single-image DPO methods perform poorly on multi-image benchmarks.
}
\label{multi-image-benchmarks}
\begin{center}
\setlength{\tabcolsep}{3pt}
\resizebox{\columnwidth}{!}{
\begin{tabular}{llllll}
\toprule
\multicolumn{1}{c}{\bf Models}  &
\multicolumn{1}{c}{\bf Parameter} &
\multicolumn{1}{c}{\bf BLINK} &
\multicolumn{1}{c}{\bf MANTIS} &
\multicolumn{1}{c}{\bf NLVR2} &
\multicolumn{1}{c}{\bf Average} \\
\midrule
GPT-4V~\citep{gpt4} & - & 51.1 & 62.7 & 88.8 & 67.53\\
\midrule
LLaVA-v1.6~\citep{LLaVAnextinterleave} & 7B & 39.6 & 45.6 & 58.9 & 48.03\\
Qwen-VL-Chat~\citep{qwenvl} & 7B & 31.2 & 39.2 & 58.7 & 43.03 \\
VideoLLaVA~\citep{videoLLaVA} & 7B & 38.9 & 35.9 & 56.5 & 43.77 \\
Fuyu~\citep{fuyu-8b} & 8B & 36.6 & 27.2 & 51.1 & 38.3 \\
Idefics2~\citep{laurençon2024matters} & 8B & 45.2 & 48.9 & 86.9 & 60.33 \\
InstructBLIP~\citep{instructblip} & 13B & 42.2 & 45.6 & 60.3 & 49.37 \\
CogVLM~\citep{wang2023cogvlm} & 17B & 41.5 & 45.2 & 58.6 & 48.43 \\
Emu2-Chat~\citep{sun2024generative} & 37B & 36.2 & 37.8 & 58.2 & 44.07 \\
\midrule
LLaVA-v1.5~\citep{liu2024improved} & 7B & 37.1 & 41.9 & 52.1 & 43.7\\
$+$ LLaVA-RLHF~\citep{sun2023aligning} & 7B & 40.8 & 30.4 & 51.8 & 41.0\\
$+$ HA-DPO~\citep{zhao2024beyond} & 7B & 38.6 & 34.6 & 51.6 & 41.6\\
$+$ POVID~\citep{zhou2024aligning} & 7B & 19.9 & 37.8 & 21.4 & 26.37\\
$+$ MIA-DPO~\citep{liu2025miadpo} & 7B & 42.9 & 44.2 & 54.2 & 47.1\\
\rowcolor[HTML]{DAEFF9}
$+$ \methodname (Ours) & 7B & \textbf{43.40} & \textbf{47.93} & \textbf{55.59} & \textbf{48.97}\\
$\Delta$ & - & \hgreen{+6.30} & \hgreen{+6.03} & \hgreen{+3.49} & \hgreen{+5.27}\\
\midrule
Qwen2.5-VL~\cite{bai2025qwen2} & 7B & 54.29 & 68.66 & 74.28 & 65.74 \\
$+$ MIA-DPO~\citep{liu2025miadpo} & 7B & 41.28 & 59.45 & 74.18 & 58.30 \\
\rowcolor[HTML]{DAEFF9}
$+$ \methodname (Ours) & 7B & \textbf{55.85} & \textbf{74.19} & \textbf{74.67} & \textbf{68.24} \\
$\Delta$ & - & \hgreen{+1.56} & \hgreen{+5.53} & \hgreen{+0.39} & \hgreen{+2.49} \\
\midrule
Qwen3-VL~\cite{bai2025qwen3} & 2B & 51.61 & 79.61 & 49.71 & 60.31 \\
\rowcolor[HTML]{DAEFF9}
$+$ \methodname (Ours) & 2B & \textbf{53.92} & \textbf{81.71} & \textbf{50.61} & \textbf{62.08} \\
$\Delta$ & - & \hgreen{+2.31} & \hgreen{+2.10} & \hgreen{+0.90} & \hgreen{+1.77} \\
\bottomrule
\end{tabular}
}
\end{center}
\end{table}
\subsection{Results on Multi-Image Benchmarks}
\noindent\textbf{Results on LLaVA-1.5:} We report the performance of our method across several multi-image benchmarks in Table~\ref{multi-image-benchmarks}. Our approach yields significant gains, specifically improving by 6.30\%, 6.03\%, and 3.49\% across three key datasets. Notably, on the complex BLINK benchmark—which requires specialized domain knowledge—\methodname outperforms the LLaVA-v1.5 baseline by 6.30\%. Furthermore, evaluations on the MANTIS and NLVR2 benchmarks show improvements of 6.03\% and 3.49\%, respectively. These results clearly highlight the effectiveness of our proposed method in enhancing the model’s capacity for visual synthesis and reasoning within multi-image contexts. We illustrate these results in Figure \ref{fig:outputs}.

\noindent\textbf{Comparison with Preference Optimization baselines}: As shown in Table~\ref{multi-image-benchmarks}, \methodname consistently outperforms established preference optimization methods. While standard LLaVA-RLHF and HA-DPO show marginal or even negative gains on specific multi-image tasks, our method achieves substantial improvements across all benchmarks. Notably, compared to the closest baseline, MIA-DPO, our approach yields an additional boost of $0.50\%,2.34\%,1.39\%$. These results indicate that our method provides a more robust signal for multi-image alignment than existing DPO or RLHF variants.

\noindent\textbf{Results on Qwen2.5-VL-7B:} As shown in Table~\ref{multi-image-benchmarks}, our method consistently improves Qwen2.5-VL-7B across all three multi-image benchmarks, yielding gains of 1.56\%, 5.53\%, and 0.39\%, with an average improvement of 2.49\%. In contrast to preference-optimization baselines such as MIA-DPO, which degrades performance on the first two benchmarks, \methodname achieves stable and substantial improvements across all settings, demonstrating its effectiveness for multi-image alignment and reasoning.

\begin{table}[t]
\caption{
\small
\textbf{Main results on single-image benchmarks.} We compare \methodname with other DPO approaches across two key single-image benchmarks. \methodname maintains strong proficiency in single-image tasks.}
\label{tab:single-image}
\begin{center}
\setlength{\tabcolsep}{4pt}
\resizebox{\columnwidth}{!}{
\begin{tabular}{lllll}
\toprule
\multicolumn{1}{c}{\bf Models} & \multicolumn{1}{c}{\bf Parameter} &
\multicolumn{1}{c}{\bf MMStar} & \multicolumn{1}{c}{\bf POPE} &
\multicolumn{1}{c}{\bf Average} \\

\midrule
LLaVA-v1.6~\citep{LLaVAnextinterleave}  & 7B & 37.6 & 70.3 & 53.95 \\
Qwen-VL-Chat~\citep{qwenvl}             & 7B & 34.5 & 74.9 & 54.7 \\
Idefics2~\citep{laurençon2024matters}   & 8B & 49.5 & 86.2 & 67.85 \\
OpenFlamingo~\citep{flaminggo}          & 9B & 36.9 & 52.6 & 44.75 \\
InstructBLIP~\citep{instructblip}       & 13B & 32.7 & 86.1 & 59.4 \\
CogVLM~\citep{wang2023cogvlm}            & 17B & 39.9 & 88.0 & 63.95 \\
Emu2-Chat~\citep{sun2024generative}     & 37B & 40.7 & 88.0 & 64.35 \\
\midrule
LLaVA-v1.5~\citep{liu2024improved}       & 7B & 32.9 & 85.9 & 59.4 \\
$+$ LLaVA-RLHF~\cite{sun2023aligning}    & 7B & 31.6 & 80.8 & 56.2 \\
$+$ HA-DPO~\citep{zhao2024beyond}        & 7B & 33.5 & 84.3 & 58.9 \\
$+$ POVID~\citep{zhou2024aligning}       & 7B & 36.2 & 86.3 & \textbf{61.25} \\
$+$ MIA-DPO~\citep{liu2025miadpo}        & 7B & 32.9 & \textbf{87.2} & 60.05 \\
\rowcolor[HTML]{DAEFF9} $+$ \methodname (Ours) & 7B & \textbf{33.62} & 85.70 & 59.66 \\
\midrule
Qwen2.5-VL \cite{bai2025qwen2}     & 7B & 62.24 & 83.13 & 72.69 \\
\rowcolor[HTML]{DAEFF9} $+$ \methodname (Ours) & 7B & \textbf{62.47} & \textbf{83.59} & \textbf{73.03} \\

\midrule
Qwen3-VL \cite{bai2025qwen3}     & 2B & 53.42 & 85.28 & 69.35 \\
\rowcolor[HTML]{DAEFF9} $+$ \methodname (Ours) & 2B & \textbf{53.62} & \textbf{85.46} & \textbf{69.54} \\
\bottomrule
\end{tabular}
}
\end{center}
\end{table}

\noindent\paragraph{Results on Qwen3-VL-2B:}
To further validate the generalizability of \methodname, we apply it to the recently released Qwen3-VL-2B~\cite{bai2025qwen3}, a compact 2B-parameter model. As shown in Tables~\ref{multi-image-benchmarks} and~\ref{tab:single-image}, \methodname consistently improves performance across both multi- and single-image benchmarks. On multi-image benchmarks, \methodname delivers consistent gains across all three evaluations, improving BLINK by $+2.31$, MANTIS by $+2.10$, and NLVR2 by $+0.90$, resulting in an overall average gain of $+1.77$ ($60.31 \rightarrow 62.08$). On single-image tasks, our method also achieves steady gains on both MMStar ($53.42 \rightarrow 53.62$) and POPE ($85.28 \rightarrow 85.46$), yielding an average improvement of $+0.19$. These results demonstrate that \methodname scales effectively to smaller, more recent architectures, achieving meaningful multi-image alignment improvements without sacrificing single-image proficiency.

\subsection{Results on Single-Image Benchmarks}
While \methodname is primarily designed to enhance performance in multi-image reasoning tasks, it is crucial that the optimized model maintains effectiveness on its core capability: single-image reasoning. We evaluate \methodname on established single-image benchmarks to verify this property. As shown in Table~\ref{tab:single-image}, \methodname consistently outperforms the LLaVA-v1.5 baseline and existing preference-alignment methods, including LLaVA-RLHF and HA-DPO, across all evaluated single-image benchmarks. These results demonstrate that \methodname not only substantially improves multi-image reasoning performance but also preserves strong single-image capabilities. This dual proficiency makes \methodname well-suited for training robust vision-language models capable of handling both single- and multi-image scenarios in real-world deployments.

\section{Ablation Studies}
\textbf{Ablation on curriculum training and data mixture: }
We study the effect of curriculum learning strategies and data mixture  on multimodal reasoning performance. Specifically, we compare incremental curriculum-based training—where models are exposed to tasks in increasing order of complexity—against flat training baselines that directly optimize on the target task distribution using different data mixtures. As summarized in Table~\ref{tab:flat}, we denote incremental setups as $A \rightarrow B$, indicating training on task set $B$ after pretraining on $A$. Across all mixtures, flat training consistently outperforms curriculum-based approaches. For example, L2 flat training attains a mean accuracy of 48.13\%, 3.3\% improvement over the L1$\rightarrow$L2 curriculum (44.83\%). This improvement corresponds to gains of 3.9\% and 5.1\% on BLINK and MANTIS, respectively. This trend generalizes across task combinations: L3 flat training improves upon L1$\rightarrow$L3 by 3.0\% (47.27\% vs.\ 44.24\%), while (L2$\cup$L3) flat training yields a 1.0\% accuracy increase over L1$\rightarrow$(L2$\cup$L3) (46.53\% vs.\ 45.56\%). Even the most gradual curriculum, L1$\rightarrow$L2$\rightarrow$L3, achieves only 45.55\% mean accuracy, underperforming all flat training variants.
\textbf{These results suggest that, for multimodal reasoning tasks, direct exposure to the target task distribution is more effective than gradual adaptation through curriculum hierarchies}. This is because curricula based training on simpler tasks induce myopic reasoning behavior, causing models to over-rely on localized or explicitly referenced visual cues present in the prompt, rather than developing a holistic, global understanding of the visual scene.

\begin{table}[t]
\centering
\caption{Effect of curriculum training strategies. Comparison of simple-to-hard (S2H) curriculum training against flat training. Flat training outperforms incremental curriculum training, as the latter leads to myopic optimization that biases the model to look for localization cues in the prompt.}
\label{tab:flat}
\setlength{\tabcolsep}{6pt}
\resizebox{\columnwidth}{!}{
\begin{tabular}{l c c c c}
\toprule
\textbf{Curriculum} & \textbf{BLINK} & \textbf{MANTIS} & \textbf{NLVR2} & \textbf{Mean} \\
\midrule
LLaVA-v1.5 & 37.1 & 41.9 & 52.1 & 43.70 \\
+L1~\citep{liu2025miadpo} & 42.9 & 44.2 & 54.2 & 47.10 \\
\rowcolor[HTML]{DAEFF9} L1$\rightarrow$ L2 & 40.2 & 39.6 & 54.7 & 44.83 \\
\rowcolor[HTML]{DAEFF9} L2 flat & 44.1 & 44.7 & 55.6 & 48.13 \\
\rowcolor[HTML]{FFE5CC} L1$\rightarrow$ L3 & 37.03 & 41.0 & 54.7 & 44.24 \\
\rowcolor[HTML]{FFE5CC} L3 flat & 43.8 & 42.8 & 55.2 & 47.27 \\
\rowcolor[HTML]{DAEFF9} L1$\rightarrow$ (L2$\cup$L3) & 40.17 & 41.0 & 55.5 & 45.56 \\
\rowcolor[HTML]{DAEFF9} (L2$\cup$L3) flat & 43.2 & 41.4 & 55.0 & 46.53 \\
L1$\rightarrow$ L2$\rightarrow$ L3 & 39.96 & 41.4 & 55.3 & 45.55 \\
\bottomrule
\end{tabular}
}
\end{table}
\begin{table}[t]
\centering
\caption{DPO vs. SFT on LLaVA-v1.5. Preference-based optimization (DPO) provides stronger guidance than standard supervised fine-tuning (SFT), leading to improved performance across benchmarks.}
\label{tab:dposft}
\setlength{\tabcolsep}{8pt}
\resizebox{0.9\columnwidth}{!}{
\begin{tabular}{l c c c c}
\toprule
\textbf{Training} & \textbf{BLINK} & \textbf{MANTIS} & \textbf{NLVR2} & \textbf{Mean} \\
\midrule
LLaVA  & 37.1 & 41.9 & 52.1 & 43.70 \\
+SFT & 40.52 & 45.15 & 55.09 & 46.92 \\
+DPO & 43.40 & 47.93 & 55.59 & 48.97 \\
\bottomrule
\end{tabular}
}
\end{table}

\noindent\textbf{Ablation on SFT vs DPO: }
We compare the effectiveness of supervised fine-tuning (SFT) and direct preference optimization (DPO) for enhancing multimodal reasoning capabilities. We simply convert the DPO data to SFT data by discarding the rejected samples. As shown in Table~\ref{tab:dposft}, both approaches substantially improve over the LLaVA-v1.5 baseline (43.70\% mean accuracy), with SFT achieving 46.92\% and DPO reaching 48.97\% mean accuracy. DPO demonstrates consistent advantages across all benchmarks, outperforming SFT by 2.88\% on BLINK (43.40\% vs. 40.52\%), 2.78\% on MANTIS (47.93\% vs. 45.15\%), and 0.41\% on NLVR2 (55.59\% vs. 55.09\%). These results indicate that preference-based optimization provides stronger learning signals for multimodal reasoning tasks compared to standard supervised learning.

\begin{figure}
    \centering
    \includegraphics[width=0.6\linewidth]{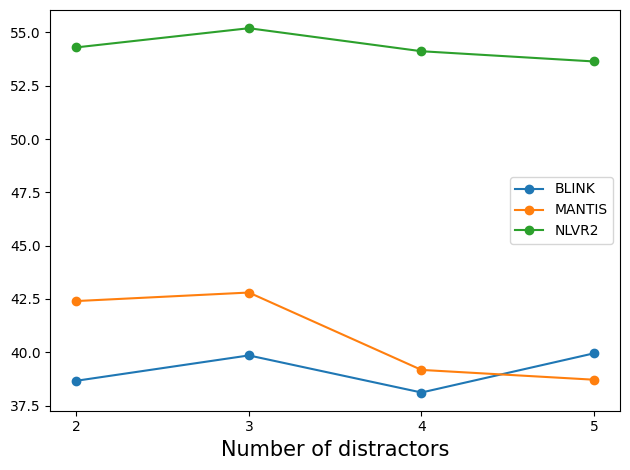}
    \caption{Effect of the number of distractors (2–5) on L3 task performance across BLINK, MANTIS, and NLVR2, showing peak accuracy at 3 distractors.}
    \label{fig:num_distractors}
\end{figure}

\noindent\textbf{Ablation on number of distractors.} We study the effect of distractor count in the L3 task by training separate models on 20K samples with 2–5 distractors. Figure~\ref{fig:num_distractors} shows the resulting accuracy trends. Performance improves as distractors increase, peaks at 3, and declines beyond that. This suggests that while more distractors enrich supervision and encourage robust reasoning, too many introduce noise. Overall, 3 distractors provide the optimal balance between contextual diversity and reasoning difficulty across benchmarks.

\noindent\textbf{Does S2H training enhance L3 ability?}
To evaluate this, we construct L3-style questions using the ImageNet validation set. For each question, we randomly sample a target image belonging to a object  $c$, and select $N \in \{2,3,4\}$ distractor images from different classes. The model is then asked: ``Which of the following images contains \texttt{object}?''
On 500 such questions, our fine-tuned model achieves an accuracy of \textbf{31.05\%}, outperforming baseline LLaVA-v1.5 by \textbf{6.94\%}. This improvement indicates that the proposed S2H training tasks enhances higher-order visual reasoning and leads to stronger L3 capability.

\begin{table}[h]
\centering
\caption{Comparison of DPO and GRPO fine-tuning on Qwen3-VL-2B across multi-image reasoning benchmarks. While GRPO achieves the highest overall performance, it exhibits instability across benchmarks.}

\setlength{\tabcolsep}{8pt}
\resizebox{0.9\columnwidth}{!}{
\begin{tabular}{lccc}
\toprule
\textbf{Method} & \textbf{MANTIS} & \textbf{BLINK} & \textbf{NLVR2} \\
\midrule
Qwen3-VL-2B & 47.00 & 79.61 & 49.71 \\
+ DPO        & 53.92 & 81.71 & 50.61 \\
+ GRPO       & 61.29 & 70.62 & 61.73 \\
\bottomrule
\end{tabular}
}
\label{tab:grpo_comparison}
\end{table}

\noindent\textbf{Effects of on-policy optimization} 
To further validate the effectiveness of our data generation strategy, we conduct additional experiments using Group Relative Policy Optimization (GRPO)~\cite{shao2024deepseekmath} as an on-policy reinforcement learning alternative to DPO. We evaluate three reward functions on Qwen3-VL-2B: (i) a \textit{length-based} reward that encourages responses close to an optimal length of 20 tokens; (ii) a \textit{format-based} reward that enforces adherence to a predefined output structure; and (iii) a \textit{rule-based} reward that measures similarity between the generated output and the gold reference answer.

As shown in Table~\ref{tab:grpo_comparison}, GRPO yields substantial gains over DPO on MANTIS and NLVR2, but exhibits a performance regression on BLINK. We attribute this instability to GRPO's  sensitivity to reward design and optimization hyperparameters, a challenge orthogonal to the primary focus of this work. Nonetheless, the strong improvements achieved with our simple reward setup further corroborate the quality of our data generation strategy for multi-image reasoning.

\section{Conclusion}

We  identify and address a critical capability gap in multi-image reasoning within Vision-Language Models. While prior methods focused heavily on localized indexing, our Simple-to-Hard (S2H) framework demonstrates that  multi-image reasoning requires a hierarchical approach—advancing from basic isolation to complex global visual search and cross-image composition. By systematically generating synthetic preference pairs across these cognitive levels, we move beyond model-specific hallucinations toward a more robust, prompt-driven alignment strategy. Our results show that this comprehensive training not only significantly improves performance on complex multi-image benchmarks  but also preserves the foundational single-image reasoning capabilities.

\section{Limitations}
Despite its effectiveness, our \methodname framework has few limitations. First, although preference pairs are generated without human annotation, the process still relies on existing datasets and a strong captioning model, whose biases may propagate into the training data. Second, our experiments focus on multi-image settings with a limited number of images (capped at 6 images); scaling to much larger image sets may introduce additional challenges in attention and efficiency that we do not explore. Finally, while \methodname  captures key forms of multi-image reasoning, it does not cover all possible reasoning patterns, such as temporal or causal reasoning across images, which we leave for future work.

\bibliography{egbib}

\begin{thebibliography}{46}
\providecommand{\natexlab}[1]{#1}

\bibitem[{Achiam et~al.(2023)Achiam, Adler, Agarwal, Ahmad, Akkaya, Aleman, Almeida, Altenschmidt, Altman, Anadkat et~al.}]{gpt4}
Josh Achiam, Steven Adler, Sandhini Agarwal, Lama Ahmad, Ilge Akkaya, Florencia~Leoni Aleman, Diogo Almeida, Janko Altenschmidt, Sam Altman, Shyamal Anadkat, and 1 others. 2023.
\newblock {GPT-4 technical report}.
\newblock \emph{arXiv preprint arXiv:2303.08774}.

\bibitem[{Awadalla et~al.(2023)Awadalla, Gao, Gardner, Hessel, Hanafy, Zhu, Marathe, Bitton, Gadre, Sagawa et~al.}]{flaminggo}
Anas Awadalla, Irena Gao, Josh Gardner, Jack Hessel, Yusuf Hanafy, Wanrong Zhu, Kalyani Marathe, Yonatan Bitton, Samir Gadre, Shiori Sagawa, and 1 others. 2023.
\newblock {Openflamingo}: An open-source framework for training large autoregressive vision-language models.
\newblock \emph{arXiv preprint arXiv:2308.01390}.

\bibitem[{Bai et~al.(2023)Bai, Bai, Yang, Wang, Tan, Wang, Lin, Zhou, and Zhou}]{qwenvl}
Jinze Bai, Shuai Bai, Shusheng Yang, Shijie Wang, Sinan Tan, Peng Wang, Junyang Lin, Chang Zhou, and Jingren Zhou. 2023.
\newblock {Qwen-VL}: A frontier large vision-language model with versatile abilities.
\newblock \emph{arXiv preprint arXiv:2308.12966}.

\bibitem[{Bai et~al.(2025{\natexlab{a}})Bai, Cai, Chen, Chen, Chen, Cheng, Deng, Ding, Gao, Ge et~al.}]{bai2025qwen3}
Shuai Bai, Yuxuan Cai, Ruizhe Chen, Keqin Chen, Xionghui Chen, Zesen Cheng, Lianghao Deng, Wei Ding, Chang Gao, Chunjiang Ge, and 1 others. 2025{\natexlab{a}}.
\newblock Qwen3-vl technical report.
\newblock \emph{arXiv preprint arXiv:2511.21631}.

\bibitem[{Bai et~al.(2025{\natexlab{b}})Bai, Chen, Liu, Wang, Ge, Song, Dang, Wang, Wang, Tang et~al.}]{bai2025qwen2}
Shuai Bai, Keqin Chen, Xuejing Liu, Jialin Wang, Wenbin Ge, Sibo Song, Kai Dang, Peng Wang, Shijie Wang, Jun Tang, and 1 others. 2025{\natexlab{b}}.
\newblock Qwen2. 5-vl technical report.
\newblock \emph{arXiv preprint arXiv:2502.13923}.

\bibitem[{Bai et~al.(2022)Bai, Jones, Ndousse, Askell, Chen, DasSarma, Drain, Fort, Ganguli, Henighan et~al.}]{bai2022training}
Yuntao Bai, Andy Jones, Kamal Ndousse, Amanda Askell, Anna Chen, Nova DasSarma, Dawn Drain, Stanislav Fort, Deep Ganguli, Tom Henighan, and 1 others. 2022.
\newblock Training a helpful and harmless assistant with reinforcement learning from human feedback.
\newblock \emph{arXiv preprint arXiv:2204.05862}.

\bibitem[{Bavishi et~al.(2023)Bavishi, Elsen, Hawthorne, Nye, Odena, Somani, and Tas{\i}rlar}]{fuyu-8b}
Rohan Bavishi, Erich Elsen, Curtis Hawthorne, Maxwell Nye, Augustus Odena, Arushi Somani, and Sagnak Tas{\i}rlar. 2023.
\newblock Introducing our multimodal models.
\newblock \emph{Adept Blog}.

\bibitem[{Betley et~al.(2025)Betley, Tan, Warncke, Sztyber-Betley, Bao, Soto, Labenz, and Evans}]{misaligned_llm}
Jan Betley, Daniel Chee~Hian Tan, Niels Warncke, Anna Sztyber-Betley, Xuchan Bao, Mart\'{\i}n Soto, Nathan Labenz, and Owain Evans. 2025.
\newblock Emergent misalignment: Narrow finetuning can produce broadly misaligned {LLM}s.
\newblock In \emph{Proceedings of the 42nd International Conference on Machine Learning}, pages 4043--4068.

\bibitem[{Chen et~al.(2024{\natexlab{a}})Chen, Li, Dong, Zhang, Zang, Chen, Duan, Wang, Qiao, Lin, and Zhao}]{mmstar}
Lin Chen, Jinsong Li, Xiaoyi Dong, Pan Zhang, Yuhang Zang, Zehui Chen, Haodong Duan, Jiaqi Wang, Yu~Qiao, Dahua Lin, and Feng Zhao. 2024{\natexlab{a}}.
\newblock Are we on the right way for evaluating large vision-language models?
\newblock In \emph{Proceedings of the 38th International Conference on Neural Information Processing Systems}.

\bibitem[{Chen et~al.(2024{\natexlab{b}})Chen, Wei, Li, Dong, Zhang, Zang, Chen, Duan, Lin, Tang, Yuan, Qiao, Lin, Zhao, and Wang}]{chen2024sharegpt4video}
Lin Chen, Xilin Wei, Jinsong Li, Xiaoyi Dong, Pan Zhang, Yuhang Zang, Zehui Chen, Haodong Duan, Bin Lin, Zhenyu Tang, Li~Yuan, Yu~Qiao, Dahua Lin, Feng Zhao, and Jiaqi Wang. 2024{\natexlab{b}}.
\newblock {ShareGPT4Video: improving video understanding and generation with better captions}.
\newblock In \emph{Proceedings of the 38th International Conference on Neural Information Processing Systems}.

\bibitem[{Chen et~al.(2023)Chen, Zhou, Zhang, Zhao, and Wen}]{chen-etal-2023-chatcot}
Zhipeng Chen, Kun Zhou, Zheng Zhang, Beichen~Gong, Xin Zhao, and Ji-Rong Wen. 2023.
\newblock {C}hat{C}o{T}: Tool-augmented chain-of-thought reasoning on chat-based large language models.
\newblock In \emph{Proceedings of the Findings of the Association for Computational Linguistics: EMNLP}.

\bibitem[{Dai et~al.(2023)Dai, Li, Li, Tiong, Zhao, Wang, Li, Fung, and Hoi}]{instructblip}
Wenliang Dai, Junnan Li, Dongxu Li, Anthony Meng~Huat Tiong, Junqi Zhao, Weisheng Wang, Boyang Li, Pascale Fung, and Steven Hoi. 2023.
\newblock Instructblip: towards general-purpose vision-language models with instruction tuning.
\newblock In \emph{Proceedings of the 37th International Conference on Neural Information Processing Systems}.

\bibitem[{Das et~al.(2026)Das, Bulat, Baldrati, Metaxas, Schiele, Tzimiropoulos, and Martinez}]{more_images}
Anurag Das, Adrian Bulat, Alberto Baldrati, Ioannis~Maniadis Metaxas, Bernt Schiele, Georgios Tzimiropoulos, and Brais Martinez. 2026.
\newblock {More Images, More Problems? A Controlled Analysis of VLM Failure Modes}.
\newblock \emph{arXiv preprint arXiv:2601.07812}.

\bibitem[{Deng et~al.(2009)Deng, Dong, Socher, Li, Li, and Fei-Fei}]{5206848}
Jia Deng, Wei Dong, Richard Socher, Li-Jia Li, Kai Li, and Li~Fei-Fei. 2009.
\newblock Imagenet: A large-scale hierarchical image database.
\newblock In \emph{Proceedings of Conference on Computer Vision and Pattern Recognition (CVPR)}, pages 248--255.

\bibitem[{Fu et~al.(2024)Fu, Hu, Li, Feng, Wang, Lin, Roth, Smith, Ma, and Krishna}]{blink}
Xingyu Fu, Yushi Hu, Bangzheng Li, Yu~Feng, Haoyu Wang, Xudong Lin, Dan Roth, Noah~A. Smith, Wei-Chiu Ma, and Ranjay Krishna. 2024.
\newblock Blink: Multimodal large language models can see but not perceive.
\newblock In \emph{Proceedings of European Conference on Computer Vision (ECCV).}

\bibitem[{Furniturewala et~al.(2024)Furniturewala, Jandial, Java, Banerjee, Shahid, Bhatia, and Jaidka}]{prompt_safety}
Shaz Furniturewala, Surgan Jandial, Abhinav Java, Pragyan Banerjee, Simra Shahid, Sumit Bhatia, and Kokil Jaidka. 2024.
\newblock ``thinking'' fair and slow: On the efficacy of structured prompts for debiasing language models.
\newblock In \emph{Proceedings of the Conference on Empirical Methods in Natural Language Processing}.

\bibitem[{Guan et~al.(2025)Guan, Roosta, Passban, and Rezagholizadeh}]{llm_sens}
Bryan Guan, Tanya Roosta, Peyman Passban, and Mehdi Rezagholizadeh. 2025.
\newblock The order effect: Investigating prompt sensitivity to input order in llms.
\newblock \emph{arXiv preprint arXiv:2502.04134}.

\bibitem[{Hurst et~al.(2024)Hurst, Lerer, Goucher, Perelman, Ramesh, Clark, Ostrow, Welihinda, Hayes, Radford et~al.}]{2024gpt4o}
Aaron Hurst, Adam Lerer, Adam~P Goucher, Adam Perelman, Aditya Ramesh, Aidan Clark, AJ~Ostrow, Akila Welihinda, Alan Hayes, Alec Radford, and 1 others. 2024.
\newblock {GPT}-4o system card.
\newblock \emph{arXiv preprint arXiv:2410.21276}.

\bibitem[{Jandial et~al.(2026)Jandial, Li, Wagle, and Koishida}]{gui_ground}
Surgan Jandial, Yinheng Li, Justin Wagle, and Kazuhito Koishida. 2026.
\newblock Do {GUI} grounders truly understand {UI} elements?
\newblock In \emph{Proceedings of Findings of the {A}ssociation for {C}omputational {L}inguistics: {EACL}}.

\bibitem[{Jandial et~al.(2025)Jandial, Wang, Bajcsy, and De~la Torre}]{fine_web}
Surgan Jandial, Yinong~Oliver Wang, Andrea Bajcsy, and Fernando De~la Torre. 2025.
\newblock On the fine-grained planning abilities of {VLM} web agents.
\newblock In \emph{Proceedings of Findings of the Association for Computational Linguistics: EMNLP}.

\bibitem[{Jiang et~al.(2024)Jiang, He, Zeng, Wei, Ku, Liu, and Chen}]{mantis}
Dongfu Jiang, Xuan He, Huaye Zeng, Cong Wei, Max~W.F. Ku, Qian Liu, and Wenhu Chen. 2024.
\newblock Mantis: Interleaved multi-image instruction tuning.
\newblock \emph{Transactions on Machine Learning Research}.

\bibitem[{Lauren{\c{c}}on et~al.(2024)Lauren{\c{c}}on, Tronchon, Cord, and Sanh}]{laurençon2024matters}
Hugo Lauren{\c{c}}on, Leo Tronchon, Matthieu Cord, and Victor Sanh. 2024.
\newblock What matters when building vision-language models?
\newblock In \emph{Proceedings of The Thirty-eighth Annual Conference on Neural Information Processing Systems}.

\bibitem[{Li et~al.(2025)Li, Zhang, Zhang, Zhang, Li, Li, MA, and Li}]{LLaVAnextinterleave}
Feng Li, Renrui Zhang, Hao Zhang, Yuanhan Zhang, Bo~Li, Wei Li, Zejun MA, and Chunyuan Li. 2025.
\newblock {LL}a{VA}-ne{XT}-interleave: Tackling multi-image, video, and {3D} in large multimodal models.
\newblock In \emph{Proceedings of The Thirteenth International Conference on Learning Representations (ICLR)}.

\bibitem[{Li et~al.(2024)Li, Xie, Li, Chen, Wang, Chen, Yang, Wang, Kong, and Liu}]{li-etal-2024-vlfeedback}
Lei Li, Zhihui Xie, Mukai Li, Shunian Chen, Peiyi Wang, Liang Chen, Yazheng Yang, Benyou Wang, Lingpeng Kong, and Qi~Liu. 2024.
\newblock {VLF}eedback: A large-scale {AI} feedback dataset for large vision-language models alignment.
\newblock In \emph{Proceedings of the Conference on Empirical Methods in Natural Language Processing}, pages 6227--6246.

\bibitem[{Li et~al.(2023)Li, Du, Zhou, Wang, Zhao, and Wen}]{pope}
Yifan Li, Yifan Du, Kun Zhou, Jinpeng Wang, Wayne~Xin Zhao, and Ji-Rong Wen. 2023.
\newblock Evaluating object hallucination in large vision-language models.
\newblock In \emph{Proceedings of The Conference on Empirical Methods in Natural Language Processing (EMNLP)}.

\bibitem[{Lin et~al.(2024)Lin, Ye, Zhu, Cui, Ning, Jin, and Yuan}]{videoLLaVA}
Bin Lin, Yang Ye, Bin Zhu, Jiaxi Cui, Munan Ning, Peng Jin, and Li~Yuan. 2024.
\newblock Video-{LL}a{VA}: Learning united visual representation by alignment before projection.
\newblock In \emph{Proceedings of the Conference on Empirical Methods in Natural Language Processing (EMNLP)}.

\bibitem[{Liu et~al.(2024{\natexlab{a}})Liu, Li, Li, and Lee}]{liu2024improved}
Haotian Liu, Chunyuan Li, Yuheng Li, and Yong~Jae Lee. 2024{\natexlab{a}}.
\newblock Improved baselines with visual instruction tuning.
\newblock In \emph{Proceedings of Computer Vision and Pattern Recognition (CVPR)}.

\bibitem[{Liu et~al.(2024{\natexlab{b}})Liu, Chu, Zang, Wei, Dong, Zhang, Liang, Xiong, Qiao, Lin et~al.}]{liu2024mmdu}
Ziyu Liu, Tao Chu, Yuhang Zang, Xilin Wei, Xiaoyi Dong, Pan Zhang, Zijian Liang, Yuanjun Xiong, Yu~Qiao, Dahua Lin, and 1 others. 2024{\natexlab{b}}.
\newblock {MMDU}: A multi-turn multi-image dialog understanding benchmark and instruction-tuning dataset for {LVLMs}.
\newblock \emph{arXiv preprint arXiv:2406.11833}.

\bibitem[{Liu et~al.(2025)Liu, Zang, Dong, Zhang, Cao, Duan, He, Xiong, Lin, and Wang}]{liu2025miadpo}
Ziyu Liu, Yuhang Zang, Xiaoyi Dong, Pan Zhang, Yuhang Cao, Haodong Duan, Conghui He, Yuanjun Xiong, Dahua Lin, and Jiaqi Wang. 2025.
\newblock {MIA}-{DPO}: Multi-image augmented direct preference optimization for large vision-language models.
\newblock In \emph{The Thirteenth International Conference on Learning Representations}.

\bibitem[{Rafailov et~al.(2024)Rafailov, Sharma, Mitchell, Manning, Ermon, and Finn}]{rafailov2024direct}
Rafael Rafailov, Archit Sharma, Eric Mitchell, Christopher~D Manning, Stefano Ermon, and Chelsea Finn. 2024.
\newblock Direct preference optimization: Your language model is secretly a reward model.
\newblock In \emph{Proceedings of the International Conference on Neural Information Processing Systems}.

\bibitem[{Rahmanzadehgervi et~al.(2025)Rahmanzadehgervi, Bolton, Taesiri, and Nguyen}]{vlm_blind}
Pooyan Rahmanzadehgervi, Logan Bolton, Mohammad~Reza Taesiri, and Anh~Totti Nguyen. 2025.
\newblock Vision language models are blind: Failing to translate detailed visual features into words.
\newblock \emph{arXiv preprint arXiv:2407.06581}.

\bibitem[{Robinson et~al.(2016)Robinson, Shao, Wu, and Fu}]{robinson2016families}
Joseph~P. Robinson, Ming Shao, Yue Wu, and Yun Fu. 2016.
\newblock Families in the wild (fiw): Large-scale kinship image database and benchmarks.
\newblock In \emph{Proceedings of the ACM on Multimedia Conference}, pages 242--246.

\bibitem[{Schulman et~al.(2017)Schulman, Wolski, Dhariwal, Radford, and Klimov}]{schulman2017proximal}
John Schulman, Filip Wolski, Prafulla Dhariwal, Alec Radford, and Oleg Klimov. 2017.
\newblock Proximal policy optimization algorithms.
\newblock \emph{arXiv preprint arXiv:1707.06347}.

\bibitem[{Shao et~al.(2024)Shao, Wang, Zhu, Xu, Song, Bi, Zhang, Zhang, Li, Wu et~al.}]{shao2024deepseekmath}
Zhihong Shao, Peiyi Wang, Qihao Zhu, Runxin Xu, Junxiao Song, Xiao Bi, Haowei Zhang, Mingchuan Zhang, YK~Li, Yang Wu, and 1 others. 2024.
\newblock Deepseekmath: Pushing the limits of mathematical reasoning in open language models.
\newblock \emph{arXiv preprint arXiv:2402.03300}.

\bibitem[{Suhr et~al.(2019)Suhr, Zhou, Zhang, Zhang, Bai, and Artzi}]{nlvr2}
Alane Suhr, Stephanie Zhou, Ally Zhang, Iris Zhang, Huajun Bai, and Yoav Artzi. 2019.
\newblock A corpus for reasoning about natural language grounded in photographs.
\newblock In \emph{Proceedings of the 57th Annual Meeting of the Association for Computational Linguistics}.

\bibitem[{Sun et~al.(2024)Sun, Cui, Zhang, Zhang, Yu, Wang, Rao, Liu, Huang, and Wang}]{sun2024generative}
Quan Sun, Yufeng Cui, Xiaosong Zhang, Fan Zhang, Qiying Yu, Yueze Wang, Yongming Rao, Jingjing Liu, Tiejun Huang, and Xinlong Wang. 2024.
\newblock Generative multimodal models are in-context learners.
\newblock In \emph{Proceedings of Computer Vision and Pattern Recognition (CVPR)}.

\bibitem[{Sun et~al.(2023)Sun, Shen, Cao, Liu, Li, Shen, Gan, Gui, Wang, Yang et~al.}]{sun2023aligning}
Zhiqing Sun, Sheng Shen, Shengcao Cao, Haotian Liu, Chunyuan Li, Yikang Shen, Chuang Gan, Liang-Yan Gui, Yu-Xiong Wang, Yiming Yang, and 1 others. 2023.
\newblock Aligning large multimodal models with factually augmented {RLHF}.
\newblock \emph{arXiv preprint arXiv:2309.14525}.

\bibitem[{Turpin et~al.(2023)Turpin, Michael, Perez, and Bowman}]{llm_faithful_cot}
Miles Turpin, Julian Michael, Ethan Perez, and Samuel~R. Bowman. 2023.
\newblock Language models don't always say what they think: Unfaithful explanations in chain-of-thought prompting.
\newblock In \emph{Thirty-seventh Conference on Neural Information Processing Systems}.

\bibitem[{Vidgen et~al.(2024)Vidgen, Agrawal, Ahmed, Akinwande, Al-Nuaimi, Alfaraj, Alhajjar, Aroyo, Bavalatti, Bartolo, Blili-Hamelin, Bollacker, Bomassani, Boston, Campos, Chakra, Chen, Coleman, Coudert, Derczynski, Dutta, Eisenberg, Ezick, Frase, Fuller, Gandikota, Gangavarapu, Gangavarapu, Gealy, Ghosh, Goel, Gohar, Goswami, Hale, Hutiri, Imperial, Jandial, Judd, Juefei-Xu, Khomh, Kailkhura, Kirk, Klyman, Knotz, Kuchnik, Kumar, Kumar, Lengerich, Li, Liao, Long, Lu, Luger, Mai, Mammen, Manyeki, McGregor, Mehta, Mohammed, Moss, Nachman, Naganna, Nikanjam, Nushi, Oala, Orr, Parrish, Patlak, Pietri, Poursabzi-Sangdeh, Presani, Puletti, Röttger, Sahay, Santos, Scherrer, Sebag, Schramowski, Shahbazi, Sharma, Shen, Sistla, Tang, Testuggine, Thangarasa, Watkins, Weiss, Welty, Wilbers, Williams, Wu, Yadav, Yang, Zeng, Zhang, Zhdanov, Zhu, Liang, Mattson, and Vanschoren}]{ai_safety}
Bertie Vidgen, Adarsh Agrawal, Ahmed~M. Ahmed, Victor Akinwande, Namir Al-Nuaimi, Najla Alfaraj, Elie Alhajjar, Lora Aroyo, Trupti Bavalatti, Max Bartolo, Borhane Blili-Hamelin, Kurt Bollacker, Rishi Bomassani, Marisa~Ferrara Boston, Siméon Campos, Kal Chakra, Canyu Chen, Cody Coleman, Zacharie~Delpierre Coudert, and 81 others. 2024.
\newblock {Introducing v0.5 of the AI Safety Benchmark from MLCommons}.
\newblock \emph{arXiv preprint arXiv:2404.12241}.

\bibitem[{Wang et~al.(2017)Wang, Robinson, and Fu}]{kinFG2017}
Shuyang Wang, Joseph~P Robinson, and Yun Fu. 2017.
\newblock Kinship verification on families in the wild with marginalized denoising metric learning.
\newblock In \emph{Proceedings of Automatic Face and Gesture Recognition (FG)}.

\bibitem[{Wang et~al.(2024)Wang, Lv, Yu, Hong, Qi, Wang, Ji, Yang, Zhao, XiXuan, Xu, Chen, Xu, Li, Dong, Ding, and Tang}]{wang2023cogvlm}
Weihan Wang, Qingsong Lv, Wenmeng Yu, Wenyi Hong, Ji~Qi, Yan Wang, Junhui Ji, Zhuoyi Yang, Lei Zhao, Song XiXuan, Jiazheng Xu, Keqin Chen, Bin Xu, Juanzi Li, Yuxiao Dong, Ming Ding, and Jie Tang. 2024.
\newblock Cog{VLM}: Visual expert for pretrained language models.
\newblock In \emph{Proceedings of The Thirty-eighth Annual Conference on Neural Information Processing Systems}.

\bibitem[{Yu et~al.(2024{\natexlab{a}})Yu, Yao, Zhang, He, Han, Cui, Hu, Liu, Zheng, Sun, and Chua}]{yu2024rlhfv}
Tianyu Yu, Yuan Yao, Haoye Zhang, Taiwen He, Yifeng Han, Ganqu Cui, Jinyi Hu, Zhiyuan Liu, Hai-Tao Zheng, Maosong Sun, and Tat-Seng Chua. 2024{\natexlab{a}}.
\newblock {RLHF-V}: Towards trustworthy {MLLMs} via behavior alignment from fine-grained correctional human feedback.
\newblock In \emph{Proceedings of Computer Vision and Pattern Recognition (CVPR)}.

\bibitem[{Yu et~al.(2024{\natexlab{b}})Yu, Zhang, Yao, Dang, Chen, Lu, Cui, He, Liu, Chua et~al.}]{yu2024rlaif}
Tianyu Yu, Haoye Zhang, Yuan Yao, Yunkai Dang, Da~Chen, Xiaoman Lu, Ganqu Cui, Taiwen He, Zhiyuan Liu, Tat-Seng Chua, and 1 others. 2024{\natexlab{b}}.
\newblock {RlHF-V}: Towards trustworthy {MLLMs} via behavior alignment from fine-grained correctional human feedback.
\newblock \emph{arXiv preprint arXiv:2405.17220}.

\bibitem[{Zhang et~al.(2024)Zhang, Dong, Zang, Cao, Qian, Chen, Guo, Duan, Wang, Ouyang et~al.}]{xcomposer2d5}
Pan Zhang, Xiaoyi Dong, Yuhang Zang, Yuhang Cao, Rui Qian, Lin Chen, Qipeng Guo, Haodong Duan, Bin Wang, Linke Ouyang, and 1 others. 2024.
\newblock {Internlm-Xcomposer-2.5}: A versatile large vision language model supporting long-contextual input and output.
\newblock \emph{arXiv preprint arXiv:2407.03320}.

\bibitem[{Zhao et~al.(2023)Zhao, Wang, Ouyang, Dong, Wang, and He}]{zhao2024beyond}
Zhiyuan Zhao, Bin Wang, Linke Ouyang, Xiaoyi Dong, Jiaqi Wang, and Conghui He. 2023.
\newblock Beyond hallucinations: Enhancing {LVLMs} through hallucination-aware direct preference optimization.
\newblock \emph{arXiv preprint arXiv:2311.16839}.

\bibitem[{Zhou et~al.(2024)Zhou, Cui, Rafailov, Finn, and Yao}]{zhou2024aligning}
Yiyang Zhou, Chenhang Cui, Rafael Rafailov, Chelsea Finn, and Huaxiu Yao. 2024.
\newblock Aligning modalities in vision large language models via preference fine-tuning.
\newblock \emph{arXiv preprint arXiv:2402.11411}.

\end{thebibliography}

\end{document}